# TODIM and TOPSIS with Z-numbers

Renato A. Krohling
Department of Production Engineering &
Graduate Program in Computer Sciences,PPGI
UFES- Federal University of Espirito Santo
Av. Fernando Ferrari, 514, CEP 29075-910 Vitória, Espirito Santo, ES, Brazil
krohling.renato@gmail.com

Guilherme Artem dos Santos &
André G. C. Pacheco
Graduate Program in Computer Sciences
UFES- Federal University of Espirito Santo
Av. Fernando Ferrari, 514, CEP 29075-910 Vitória, Espirito Santo, ES, Brazil
guilherme.artem@gmail.com
pacheco.comp@gmail.com

**Abstract**

In this paper, we present an approach that is able to handle with Z-numbers in the context of Multi-Criteria Decision Making (MCDM) problems. Z-numbers are composed of two parts, the first one is a restriction on the values that can be assumed, and the second part is the reliability of the information. As human beings we communicate with other people by means of natural language using sentences like: the journey time from home to university takes *about half hour, very likely*. Firstly, Z-numbers are converted to fuzzy numbers using a standard procedure. Next, the Z-TODIM and Z-TOPSIS are presented as a direct extension of the fuzzy TODIM and fuzzy TOPSIS, respectively. The proposed methods are applied to two case studies and compared with the standard approach using crisp values. Results obtained show the feasibility of the approach. In addition, a graphical interface was built to handle with both methods Z- TODIM and Z-TOPSIS allowing ease of use for user in other areas of knowledge.

## 1. Introduction

The theory of fuzzy sets and fuzzy logic developed by Zadeh (1965) has been applied to model uncertainty or lack of knowledge when applied to a variety of MCDM problems. Bellman & Zadeh (1970) introduced the theory of fuzzy sets in problems of MCDM as an effective approach to treat vagueness, lack of knowledge and ambiguity inherent in the human decision making process, which are known as fuzzy multi-criteria decision making (FMCDM).

For real world-problems the decision matrix is affected by uncertainty, which may be modeled using fuzzy numbers. A fuzzy number (Dubois and Prade, 1980) can be seen as an extension of an interval with varied grade of membership. This means that each value in the interval has associated a real number that indicates its compatibility with the vague statement associated with a fuzzy number. MCDM methods like the standard TOPSIS (Hwang and Yoon, 1980) and TODIM (Gomes and Rangel, 1992) in the last few years have largely been applied to a variety of MCDM problems.

TOPSIS has been generalized to deal with a variety of information types. For example, interval numbers (Jahanshahloo, Lotfi, & Davoodi, 2009; Dymova, Sevastjanov, Tikhonenko, 2013), probability distributions (Wentao, Huan, 2010), fuzzy information (Chen, 2000; Chen, Tsao, 2003; Wang, Liu, and Zhang, 2005; Wang, Lee, and Lin, 2003; Wang, and Lee, 2008, Krohling and Campanharo, 2011), intuitionistic fuzzy information (Yue, 2014), interval-valued intuitionistic fuzzy information (Ye, 2010, Park et al., 2011), among many others. For the reader interested in TOPSIS for hybrid data types, we recommend Lourenzutti and Krohling (2016).

The TODIM method has also been extended to different types of information. Firstly, Krohling and de Souza (2012) proposed the fuzzy TODIM to treat fuzzy numbers. Another important aspect in the treatment of uncertainties is the reliability of information. So, the fuzzy TODIM was extended to intuitionistic fuzzy TODIM (Krohling, Pacheco and Siviero, 2013; Lourenzutti and Krohling, 2013). The latter is able to deal with uncertainty described by intuitionistic fuzzy numbers (Atanassov, 1986). Zhang, and Xu (2014) extended the TODIM for hesitant fuzzy information. For information described by probability distributions were developed the Hellinger TODIM and the Hellinger TOPSIS (Lourenzutti and Krohling, 2014; Krohling, Lourenzutti, and Campos, 2015). However, in many decision-making problems, the data come from different sources, which means that data can be presented in numerical format, interval, fuzzy, fuzzy intuitionist, or even described by probability distributions. So, a method for decision-making should be able to process simultaneously hybrid information, i.e., information in different formats is very desired and have been presented (Wang and Wang, 2008; Fan et al. (2013); Lourenzutti and Krohling 2016).

Both methods TOPSIS and TODIM and their extension have successfully been applied to tackle different kinds of information to solve MCDM problems, but one very important aspect in decision making is the reliability of the information involved in the process. Z-numbers were proposed by Zadeh (2011) as a new way to treat uncertainty and reliability of information. Z-numbers are composed of two parts, the first one is a restriction on the values that can be assumed, and the second part is the reliability of the information. Z-numbers can be used to model sentences like "the temperature in summer will be very hot, very sure," or "the journey time from home to university takes about half hour, very likely." As human beings we communicate with other people by means of natural language using sentences like the ones mentioned above. The standard approach to work with Z-number was presented by Kang et al. (2012a), where a procedure to convert Z-numbers into fuzzy numbers has been proposed. The same authors, Kang et al. (2012b) apply the procedure to solve MCDM problems after converting Z-numbers to crisp values.

Azadeh et al. (2013) presented a new AHP method based on Z-number to deal with linguistic decision making problems. An application to search the criteria's for the evaluation of best universities. Xiao (2014) proposed a method for fuzzy multi-criteria decision making with Z-numbers, where the evaluation of each alternative with respect to each criterion is described as a Z-number. The Z-numbers in the decision making matrix are converted to crisp numbers and the author show the effectiveness of their method. Patel, Rahimi, & Khorasani (2015) provided an applied model of Z-numbers based on certain realistic assumptions regarding probability distributions. Aliev, Alizadeh, Huseynov (2015) proposed an arithmetic to discrete Z-numbers as addition, subtraction, multiplication, division, square root and other operations. The authors state that problems involving computation with Z-numbers is far from easy to solve.

In this paper, the goal consists in the extension of fuzzy TODIM to handle uncertain decision matrices described by Z-numbers. In section 2, we provide some background on fuzzy multicriteria decision making. In addition we review fuzzy TODIM and fuzzy TOPSIS. In section 3, we give the necessary background on Z-numbers, and the procedure to convert Z-numbers into trapezoidal fuzzy numbers. Next, we present the Z-TODIM and the Z-TOPSIS. In section 4, we illustrate the method using two case studies showing the results and the feasibility of the approach. In section 5, we present some conclusions and directions for future works.

## 2 Fuzzy Multi-criteria Decision Making

In this section, we provide some basic knowledge of fuzzy sets and fuzzy numbers (Wang & Lee, 2009), (Sanayei, Mousavi, and Yazdankhah, 2010).

### 2.1 Preliminaries on fuzzy numbers

**Definition 1:** A fuzzy set $\tilde{A}$ in a universe of discourse $X$ is characterized by a membership function $\mu_{\tilde{A}}(x)$ that assigns each element $x$ in $X$ a real number in the interval [0; 1]. The numeric value $\mu_{\tilde{A}}(x)$ stands for the grade of membership of $x$ in $\tilde{A}$.

**Definition 2:** A trapezoidal fuzzy number $\tilde{a}$ is defined by a quadruplet $\tilde{a} = (a_1, a_2, a_3, a_4)$ with membership function given by:

$$\mu_{\tilde{a}}(x) = \begin{cases} 0, & x < a_1 \\ \dfrac{x - a_1}{a_2 - a_1}, & a_2 \geq x \geq a_1 \\ 1, & a_2 \geq x \geq a_3 \\ \dfrac{x - a_3}{a_3 - a_4}, & a_3 \geq x \geq a_4 \\ 0, & x > a_4 \end{cases}$$

**Definition 3:** Let a trapezoidal fuzzy number $\tilde{a} = (a_1, a_2, a_3, a_4)$, then the defuzzified value $m(\tilde{a})$ is calculated by:

$$m(\tilde{a}) = \frac{(a_1 + a_2 + a_3 + a_4)}{4}. \tag{5}$$

**Definition 5:** Let two trapezoidal fuzzy numbers $\tilde{a} = (a_1, a_2, a_3, a_4)$ and $\tilde{b} = (b_1, b_2, b_3, b_4)$, then the distance between them is calculated as (Mahdavi *et al.*, 2008):

$$d(\tilde{a}, \tilde{b}) = \sqrt{\frac{1}{6}\left[\sum_{i=1}^{4}(b_i - a_i)^2 + \sum_{i \in \{1,3\}}(b_i - a_i)(b_{i+1} - a_{i+1})\right]}. \tag{6}$$

## 2.2 Decision matrix described by fuzzy numbers

Let us consider the fuzzy decision matrix *A*, which consists of a*lternatives* and *criteria*, described by:

$$A = \begin{matrix} & \begin{matrix} C_1 & \cdots & C_n \end{matrix} \\ \begin{matrix} A_1 \\ \cdots \\ A_m \end{matrix} & \begin{pmatrix} \tilde{x}_{11} & \cdots & \tilde{x}_{1n} \\ \vdots & \ddots & \vdots \\ \tilde{x}_{m1} & \cdots & \tilde{x}_{mn} \end{pmatrix} \end{matrix}$$

where $A_1, A_2, \cdots, A_m$ are alternatives, $C_1, C_2, \ldots, C_n$ are criteria, $\tilde{x}_{ij}$ are fuzzy numbers that indicates the rating of the alternative $A_i$ with respect to criterion $C_j$. The weight vector $W = (w_1, w_2, \ldots, w_n)$ composed of the individual weights $w_j (j=1,\ldots,n)$ for each criterion $C_j$ satisfying $\sum_{j=1}^{n} w_j = 1$.

## 2.3 Fuzzy TODIM

The Fuzzy TODIM method (Krohling and de Souza, 2012), is described in the following steps:

**Step 1:** The criteria are normally classified into two types: *benefit* and *cost*. The fuzzy-decision matrix $\tilde{A} = [\tilde{x}_{ij}]_{mxn}$ with $i = 1, \ldots, m$, and $j = 1, \ldots, n$ is normalized that results the correspondent fuzzy decision matrix $\tilde{R} = [\tilde{r}_{ij}]_{mxn}$. The fuzzy normalized value $\tilde{r}_{ij}$ is calculated as:

$$r_{ij}^k = \frac{\max(a_{ij}^4) - a_{ij}^k}{\max_i(a_{ij}^4) - \min_i a_{ij}^1} \quad \text{with } k=1,2,3,4 \qquad \text{for cost criteria}$$

$$r_{ij}^k = \frac{a_{ij}^k - \min(a_{ij}^1)}{\max_i(a_{ij}^4) - \min_i a_{ij}^1} \quad \text{with } k=1,2,3,4 \qquad \text{for benefit criteria} \qquad (7)$$

**Step 2:** Calculate the dominance of each alternative $\tilde{A}_i$ over each alternative $\tilde{A}_j$ using the following expression:

$$\delta(\tilde{A}_i, \tilde{A}_j) = \sum_{c=1}^m \phi_c(\tilde{A}_i, \tilde{A}_j) \qquad \forall (i,j) \qquad (8)$$

where

$$\phi_c(\tilde{A}_i, \tilde{A}_j) = \begin{cases} \sqrt{w_c \cdot d(\tilde{x}_{ic}, \tilde{x}_{jc})} & \text{if } [m(\tilde{x}_{ic}) - m(\tilde{x}_{jc})] > 0 \\ 0, & \text{if } [m(\tilde{x}_{ic}) - m(\tilde{x}_{jc})] = 0 \\ \frac{-1}{\theta}\sqrt{w_c \cdot d(\tilde{x}_{ic}, \tilde{x}_{jc})} & \text{if } [m(\tilde{x}_{ic}) - m(\tilde{x}_{jc})] < 0 \end{cases} \qquad (9)$$

The term $\phi_c(\tilde{A}_i, \tilde{A}_j)$ represents the contribution of the criterion $c$ to the function $\delta(\tilde{A}_i, \tilde{A}_j)$ when comparing the alternative $i$ with alternative $j$. The parameter $\theta$ represents the attenuation factor of the losses, which can be tuned according to the problem at hand. In Expression (9) $m(\tilde{x}_{ic})$ and $m(\tilde{x}_{jc})$ stands for the defuzzified values of the fuzzy number $\tilde{x}_{ic}$ and $\tilde{x}_{jc}$, respectively. The term $d(\tilde{x}_{ic}, \tilde{x}_{jc})$ designates the distance between the two fuzzy numbers $\tilde{x}_{ic}$ and $\tilde{x}_{jc}$, as defined in Eq. (6). Three cases can occur in Eq. (9): i) if the value $m(\tilde{x}_{ic}) - m(\tilde{x}_{jc})$ is positive, it represents a gain; ii) if the value $m(\tilde{x}_{ic}) - m(\tilde{x}_{jc})$ is nil; and iii) if the value $m(\tilde{x}_{ic}) - m(\tilde{x}_{jc})$ is negative, it represent a loss. The final matrix of dominance is obtained by summing up the partial matrices of dominance for each criterion.

**Step 2:** Calculate the global value of the alternative $i$ by normalizing the final matrix of dominance according to the following expression:

$$\xi_i = \frac{\sum \delta(i,j) - \min \sum \delta(i,j)}{\max \sum \delta(i,j) - \min \sum \delta(i,j)} \qquad (10)$$

Ordering the values $\xi_i$ provides the rank of each alternative. The best alternatives are those that have higher value $\xi_i$.

### 2.4 Fuzzy TOPSIS

Next, we describe the fuzzy TOPSIS, since we will apply it as well. In this case, the weighted normalized fuzzy-decision matrix $\tilde{P} = [\tilde{p}_{ij}]_{mxn}$ with $i = 1,...,m$, and $j = 1,...,n$ is constructed by multiplying the normalized fuzzy-decision matrix by its associated weights. The weighted fuzzy normalized value $\tilde{p}_{ij}$ is calculated as:

$$\tilde{p}_{ij} = w_i \cdot \tilde{p}_{ij} \text{ with } i = 1,...,m, \text{ and } j = 1,...,n. \qquad (10)$$

The fuzzy TOPSIS is described as follows.

**Step 1:** Identify the positive ideal solution $A^+$ (benefits) and negative ideal solution $A^-$ (costs) as follows:

$$A^+ = (\tilde{p}_1^+, \tilde{p}_2^+, ..., \tilde{p}_m^+) \quad (11)$$

$$A^- = (\tilde{p}_1^-, \tilde{p}_2^-, ..., \tilde{p}_m^-) \quad (12)$$

where

$$\tilde{p}_j^+ = \left(\max_i \tilde{p}_{ij}, j \in J_1; \min_i \tilde{p}_{ij}, j \in J_2\right) \quad (13)$$

$$\tilde{p}_j^- = \left(\min_i \tilde{p}_{ij}, j \in J_1; \max_i \tilde{p}_{ij}, j \in J_2\right) \quad (14)$$

where $J_1$ and $J_2$ represent the criteria *benefit* and *cost*, respectively.

**Step 2:** Calculate the Euclidean distances from the positive ideal solution $A^+$ and the negative ideal solution $A^-$ of each alternative $A_i$, respectively as follows:

$$d_i^+ = \sum_{j=1}^n d\left(\tilde{p}_{ij}, \tilde{p}_j^+\right), \quad \text{with } i = 1,...,m. \quad (15)$$

$$d_i^- = \sum_{j=1}^n d\left(\tilde{p}_{ij}, \tilde{p}_j^-\right), \quad \text{with } i = 1,...,m. \quad (16)$$

where the distance $d\left(\tilde{p}_{ij}, \tilde{p}_j^+\right)$ between two fuzzy numbers is defined according to Equation 9.

**Step 3:** Calculate the relative closeness $\xi_i$ for each alternative $A_i$ with respect to positive ideal solution as given by

$$\xi_i = \frac{d_i^-}{d_i^+ + d_i^-}. \quad (17)$$

**Step 4:** Rank the alternatives according to the relative closeness. The best alternatives are those that have higher value $\xi_i$ and therefore should be chosen because they are closer to the positive ideal solution.

### 3. Decision making with Z-numbers
In this section, we provide some basic knowledge of Z-numbers (Zadeh, 2011; Kang et al., 2012). 2010).

### 3.1 Preliminaries on Z-number
Zadeh (2011) introduced the concept of a Z-number and methods of computation with Z-numbers have been outlined. A Z-number is defined by the tuple (A, B), where A and B are fuzzy numbers, and a Z-valuation by a tuple of the form (X, A, B), which can be understood by X is (A, B), where X is a variable. Z-valuations can be used to model sentences like the "temperature in summer will be very hot, very sure",  whereas X is the variable that is "the temperature", A is  the fuzzy number "very hot" and B the number "very sure". In the above example you can get a better idea of the meaning of A and B by interpreting  A as a restriction of the values that the temperature can assume, i.e., very hot and B as a fuzzy restriction on the probability of X be A, i.e., (Prob (X is A) is B). Since Z-numbers represent a relatively new concept, some operations as distance calculation between two Z-numbers are not defined yet. So, in order to overcome the difficulties to work with Z numbers is required a conversion to standard fuzzy numbers.

## 3.2 Conversion of Z-number in fuzzy number

Kang et al. (2012) defined an operation for converting Z numbers to fuzzy numbers by means of the calculation of the fuzzy expectation. This method can be summarized in two steps:

**Step 1:** Given a Z-number (A, B), firstly the reliability B is transformed in a crisp number using centroid method which calculated by:

$$\alpha = \frac{\int x\mu_B(x)dx}{\int \mu_B(x)dx} \quad (18)$$

For trapezoidal fuzzy numbers, the centroid is calculated by:

$$\alpha = \frac{b_1 + b_2 + b_3 + b_4}{4} \quad (19)$$

**Step 2:** Calculate the fuzzy number Z' from the Z-number (A, B) by:

$$Z' = \left\{ \langle x, \mu_{Z'}(x) \rangle \mid \mu_{Z'}(x) = \mu_A\left(\frac{x}{\sqrt{\alpha}}\right) \right\} \quad (20)$$

If A is a trapezoidal fuzzy number, then Z´ is calculated by:

$$Z' = \left(\sqrt{\alpha}a_1, \sqrt{\alpha}a_2, \sqrt{\alpha}a_3, \sqrt{\alpha}a_4\right) \quad (21)$$

Next, we describe the Z-TODIM.

## 3.3 Z-TODIM

The method can be summarized in the following steps:

**Step 1:** Define the decision matrix in terms of Z-numbers X(A, B).

**Step 2:** Convert the decision matrix with Z-numbers X(A, B) into the correspondent fuzzy number (Z´).

**Step 3:** Apply the Fuzzy TODIM.

In case of the Z-TOPSIS, the only change necessary is to replace the application of Fuzzy TODIM by Fuzzy TOPSIS in step 3.

In the following, we illustrate the approach by means of two case studies.

## 4. Experimental Results

### 4.1. Case study 1 - vehicle choice

This case study presented by Kang et al. (2012) consists in the selection of a vehicle for a journey among three possible choices: car, taxi and train. The criteria are: price, journey time and comfort. Price and journey time are cost criteria and comfort is a benefit criterion. In this study, both the ratings as well as the criteria weights are described by Z-numbers as shown in Table 1.

The linguistic values VH, H, and M stands for very High, High and Medium, respectively. After conversion of the linguistic values into Z numbers expressed with numeric values is shown in Table 2.

Table 1. Decision matrix described by Z-numbers expressed with linguistic values.

|  | Price (Pounds) $C_1$ (VH,VH) | Journey Time (min) $C_2$ (H,VH) | Comfort $C_3$ (M,VH) |
|---|---|---|---|
| $A_1$ (Car) | ((9,10,12),VH) | ((70,100,120),M) | ((4,5,6),H) |
| $A_2$ (Taxi) | ((20,24,25),H) | ((60,70,100),VH) | ((7,8,10),H) |
| $A_3$ (Train) | ((15,15,15),H) | ((70,80,90),H) | ((1,4,7),H) |

Table 2. Decision matrix described by Z-numbers (Kang et al., 2012).

|  | Price (Pounds) $C_1$ ((0.75,1,1),(0.75,1,1)) | Journey Time (min) $C_2$ ((0.5,0.75,1),(0.75,1,1)) | Comfort $C_3$ ((0.25,0.5,0.75),(0.75,1,1) |
|---|---|---|---|
| $A_1$ (Car) | ((9,10,12),(0.75,1,1)) | ((70,100,120),(0.75,1,1)) | ((4,5,6),(0.5,0.75,1)) |
| $A_2$ (Taxi) | ((20,24,25),(0.5,0.75,1)) | ((60,70,100),(0.75,1,1)) | ((7,8,10),(0.5,0.75,1)) |
| $A_3$ (Train) | ((15,15,15),(0.5,0.75,1)) | ((70,80,90),(0.5,0.75,1)) | ((1,4,7),(0.5,0.75,1)) |

Table 3. Decision matrix after conversion to *fuzzy* number.

|  | Price (Pounds) $C_1$ (0.72, 0.96, 0.96) | Journey Time (min) $C_2$ ((0.5,0.75,1),(0.75,1,1)) | Comfort $C_3$ ((0.25,0.5,0.75),(0.75,1,1)) |
|---|---|---|---|
| $A_1$ (Car) | (8.62, 9.57, 11.49) | (49.50, 70.71, 84.85) | (3.46, 4.33, 5.20) |
| $A_2$ (Taxi) | (17.32, 20.79, 21.65) | (57.45, 67.02, 95.75) | (6.06, 6.93, 8.660) |
| $A_3$ (Train) | (12.99, 12.99, 12.99) | (60.62, 69.28, 77.94) | (0.87, 3.46, 6.06) |

The Z-TODIM and Z-TOPSIS were applied to this problem after converting the Z-number to trapezoidal fuzzy numbers as shown in Table 4. The fuzzy weights for the three criteria were converted to crisp values (Kang et al., 2012). In table 4, we present the results obtained by Z-TODIM, Z-TOPSIS and standard TOPSIS with crisp values provided by Kang et al. (2012). The final ranking is also illustrated in Figure 1. We can note that the alternative $A_1$ (Car) is the best alternative, followed by $A_3$ (train) and the worst is $A_2$ (Taxi) obtained by Z-TODIM as well as by Z-TOPSIS. We observe that in the approach by Kang et al. (2012) using TOPSIS with crisp values (simplified version) occur a change in the order of the alternatives $A_2$ (taxi) with alternative $A_3$ (train). In order to investigate the influence of $\theta$, we carry out a sensitivity study simulating the Z-TODIM for several values of $\theta$ {0.5, 0.8, 1.0, 1.2, 1.5, 1.8, 2.0, 2.5, 5} and the ranking is shown in Table 5. We can notice that the final ranking did not change with $\theta$.

Table 4. Ranking of the alternatives for case study 1.

| Alternatives | Z-TODIM | Z-TOPSIS | TOPSIS (Kang et al., 2012) |
|---|---|---|---|
| $A_1$ (Car) | 1.000 | 0.2305 | 0.36 |
| $A_2$ (Taxi) | 0 | 0.1363 | 0.29 |
| $A_3$ (Train) | 0.2388 | 0.1856 | 0.28 |

Table 5. Sensitivity study for case study 1 regarding $\theta$.

| Alternatives | order | $\theta = 0.5$ | $\theta = 0.8$ | $\theta = 1$ | $\theta = 1.2$ | $\theta = 1.5$ | $\theta = 1.8$ | $\theta = 2.5$ | $\theta = 5$ |
|---|---|---|---|---|---|---|---|---|---|
| $A_1$ (Car) | 1 | 1.0 | 1.0 | 1.0 | 1.0 | 1.0 | 1.0 | 1.0 | 1.0 |
| $A_2$ (Taxi) | 3 | 0 | 0 | 0 | 0 | 0 | 0 | 0 | 0 |
| $A_3$ (Train) | 2 | 0.3012 | 0.2722 | 0.2570 | 0.2442 | 0.2283 | 0.2155 | 0.1933 | 0.1540 |

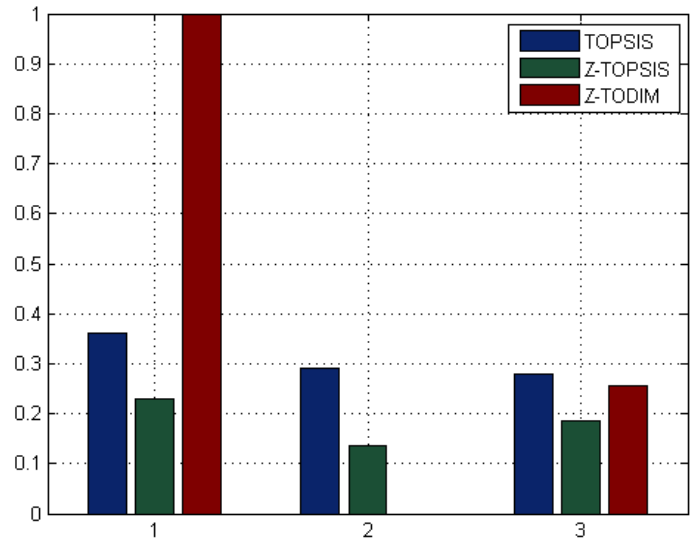

Figure 1: Final ranking of the alternatives for case study 1.

### 4.2. Case study 2 – clothing evaluation

This case study proposed by Xiao (2014) consists in the evaluation of certain type of clothing by male customers. The alternatives are: like $A_1$ (very much), $A_2$ (like), $A_3$ (ordinary) and $A_4$ (dislike). The criteria are: color and style $C_1$, durability $C_2$ and price $C_3$. All the criteria are benefits. The ratings of the alternatives are described by Z-numbers as shown in Table 6. The linguistic values VS, S, and NVS stands for Very Sure, Sure and Not Very Sure, respectively. After conversion of the linguistic values into Z-numbers expressed with numeric values is shown in Table 7. The weights of criteria are crisp values.

Table 6. Decision matrix described by Z-numbers

|  | $C_1$ 0.35 | $C_2$ 0.5 | $C_3$ 0.15 |
|---|---|---|---|
| $A_1$ | (about 20%, VS) | (nearly 10%, S) | (about 10%, VS) |
| $A_2$ | (nearly 40%, S) | (nearly 60%, S) | (over 40%, NVS) |
| $A_3$ | (about 30%, NVS) | (about 20%, S) | (over 30%, VS) |
| $A_4$ | (about 1%, VS) | (about 10%, VS) | (about 20%, S) |

Table 7. Decision matrix described by Z-numbers expressed as numeric values (Xiao, 2014).

|  | $C_1$ 0.35 | $C_2$ 0.5 | $C_3$ 0.15 |
|---|---|---|---|
| $A_1$ | (0,0.15,0.25,0.35),(0,0.2,0.35;0.8)) | ((0,0.03,0.12,0.2),(0,0.1,0.2;0.8)) | ((0,0.08,0.16,0.2),(0,0.1,0.2;0.9)) |
| $A_2$ | ((0.25,0.35,0.42,0.5),(0.3,0.4,0.5;0.8)) | ((0.4,0.5,0.65,0.75),(0.4,0.6,0.75;0.8)) | ((0.3,0.35,0.45,0.55),(0.3,0.4,0.55;0.7)) |
| $A_3$ | ((0.2,0.25,0.35,0.45),(0.2,0.3,0.45;0.7)) | ((0.1,0.15,0.25,0.35),(0.1,0.2,0.35;0.9)) | ((0.25,0.3,0.38,0.45),(0.25,0.3,0.45;0.9)) |
| $A_4$ | ((0,0.08,0.1,0.2), (0,0.1,0.2;0.8)) | ((0,0.07,0.16,0.2), (0,0.1,0.2;0.8)) | ((0.1,0.15,0.25,0.35), (0.1,0.2,0.35;0.9)) |

Table 8. Decision matrix after conversion to *fuzzy* number.

|  | $C_1$ 0.35 | $C_2$ 0.5 | $C_3$ 0.15 |
|---|---|---|---|
| $A_1$ | (0, 0.15, 0.24, 0.34) | (0, 0.03, 0.10, 0.17) | (0, 0.08, 0.15, 0.19) |
| $A_2$ | (0.22, 0.30, 0.36, 0.43) | (0.34, 0.43, 0.56, 0.65) | (0.21, 0.25, 0.32, 0.39) |
| $A_3$ | (0.14, 0.18, 0.25, 0.32) | (0.09, 0.13, 0.22, 0.30) | (0.24, 0.29, 0.37, 0.44) |
| $A_4$ | (0, 0.08, 0.10, 0.19) | (0, 0.07, 0.15, 0.19) | (0.09, 0.13, 0.22, 0.30) |

The Z-TODIM and Z-TOPSIS were applied to this problem after converting the Z-number to trapezoidal fuzzy numbers (Xiao, 2014). In table 9, we present the results obtained by Z-TODIM, Z-TOPSIS and those provided by Xiao (2014). The final ranking is also illustrated in Figure 2. We can note that the alternative $A_2$ (like) is the best alternative, followed by $A_3$ (ordinary) and the worst are $A_4$ (dislike) and $A_1$ (very likely) obtained by Z-TODIM. The results are very similar as compared with those obtained by Z-TOPSIS and Xiao (2014), whereas the small differences lie in the worst alternatives $A_1$ and $A_4$. In order to investigate the influence of $\theta$, we carry out a sensitivity study simulating the Z-TODIM for several values of $\theta$ {0.5, 0.8, 1.0, 1.2, 1.5, 1.8, 2.0, 2.5, 5} and the ranking is shown in Table 10. We can notice that the final ranking did not change with $\theta$.

Table 9. Ranking of the alternatives for case study 2.

| Alternatives | Z-TODIM | Z-TOPSIS | (Xiao, 2014) |
|---|---|---|---|
| $A_1$ | 0 | 0.0429 | 0.1362 |
| $A_2$ | 1.0 | 0.2539 | 0.4821 |
| $A_3$ | 0.4942 | 0.1207 | 0.2629 |
| $A_4$ | 0.0451 | 0.0348 | 0.1242 |

Table 10. Sensitivity study for case study 2 regarding $\theta$.

| Alternatives | order | $\theta = 0.5$ | $\theta = 0.8$ | $\theta = 1$ | $\theta = 1.2$ | $\theta = 1.5$ | $\theta = 1.8$ | $\theta = 2.5$ | $\theta = 5$ |
|---|---|---|---|---|---|---|---|---|---|
| $A_1$ | 4 | 0 | 0 | 0 | 0 | 0 | 0 | 0 | 0 |
| $A_2$ | 1 | 1 | 1.0 | 1.0 | 1.0 | 1.0 | 1.0 | 1.0 | 1.0 |
| $A_3$ | 2 | 0.6095 | 0.5872 | 0.5766 | 0.5682 | 0.5584 | 0.5509 | 0.5388 | 0.5197 |
| $A_4$ | 3 | 0.0058 | 0.0075 | 0.0084 | 0.0090 | 0.0098 | 0.0103 | 0.0113 | 0.0127 |

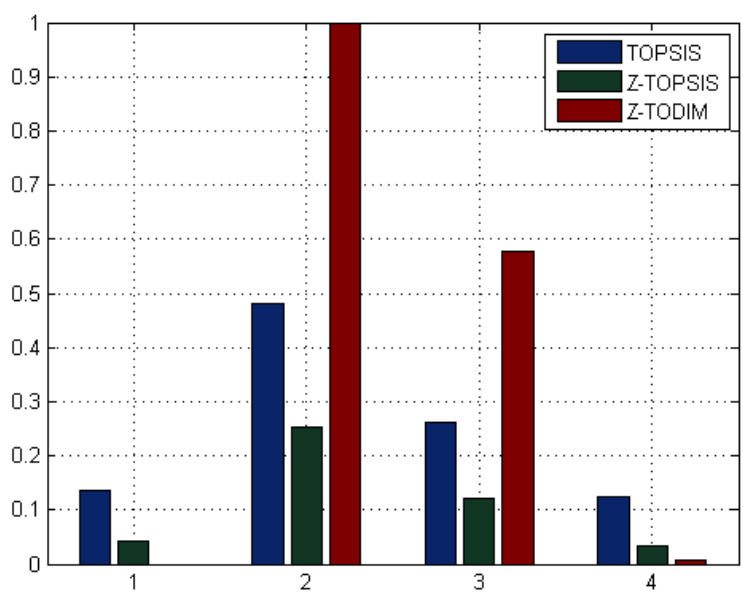

Figure 2: Final ranking of the alternatives for case study 2.

## 5. Conclusions

In this work, we present a Z-TODIM and Z-TOPSIS, which is able to tackle MCDM problems modeled by Z-numbers. The Z-TODIM method is based on the conversion of Z-numbers into trapezoidal fuzzy numbers and in turn, the application of the fuzzy TODIM or fuzzy TOPSIS. The methods have been applied to two case studies with promising results. The rank of the alternatives obtained by Z-TODIM compared to Z-TOPSIS are almost the same. Since there is no closed expression to calculate the distance between Z-numbers, our approach turns out to a promising way. We currently are expanding the Z-TODIM and Z-TOPSIS to be applied to other challenging MCDM problems modeled by Z-numbers.

**Acknowledgements:** R.A. Krohling would like to thank the financial support of the Brazilian agency CNPq under grant nr. 309161/2015-0.

## References

Aliev, R.A., Alizadeh, A.V., Huseynov, O.H. The arithmetic of discrete Z-numbers**,** Information Sciences, 290 (2015), 134-155.

Atanassov. K.T. Intuitionistic fuzzy sets, Fuzzy Sets and Systems, 20 (1986), 87-96.

Azadeh, A., Saberi, M., Atashbar, N.Z., Chang, E., Pazhoheshfar, P. Z-AHP: A Z-number extension of fuzzy analytical hierarchy process. In 7th IEEE International Conference on Digital Ecosystems and Technologies (DEST), 2013.

Chen, C.T. Extensions of the TOPSIS for group decision-making under fuzzy environment, Fuzzy Sets and Systems 114 (2000) 1-9.

Chen, T.Y., Tsao, C.Y. The interval valued fuzzy TOPSIS methods and experimental analysis, Fuzzy Sets and System 11 (2008) 1410-1428.

Dubois, D., Prade, H. Fuzzy Sets and Systems: Theory and Applications. New York: Academic Press, 1980.

Dymova, L., Sevastjanov, P. Tikhonenko, A. A direct interval extension of TOPSIS method, Expert Systems with Applications 40(12) (2013) 4841- 4847.

Fan, Z.-P., Zhang, X., Chen, F,-D., Yang L. Extended TODIM method for hybrid multiple attribute decision making problems, Knowledge-Based Systems, 42 (2013), 40-48.

Gomes, L.F.A.M., & Lima, M.M.P.P. TODIM: Basics and application to multicriteria ranking of projects with environmental impacts. Foundations of Computing and Decision Sciences, 16 (4) (1992), 113-127.

Gomes, L.F.A.M., & Rangel, L.A.D. An application of the TODIM method to the multicriteria rental evaluation of residential properties. European Journal of Operational Research, 193 (2009), 204-211.

Hwang, C.L. & Yoon, K.P. Multiple attributes decision making methods and applications. Berlin: Springer-Verlag, 1981.

Jahanshahloo, G., Lotfi, F.H., Davoodi, A. Extension of TOPSIS for decision-making problems with interval data: Interval efficiency, Mathematical and Computer Modelling 49(5–6) (2009) 1137-1142.

Kang, B., Wei, D., Li, Y. and Deng, Y. A method of converting Z-number to classical fuzzy number, Journal of Information and Computational Science, 9(3) (2012b), 703–709.

Kang, B., Wei, D., Li, Y. and Deng, Y. Decision Making Using Z-numbers under Uncertain Environment, Journal of Computational Information Systems, 8(7) (2012a), 2807-2814.

Krohling, R.A., & de Souza, T.T M. Combining prospect theory and fuzzy numbers to multi- criteria decision making. Expert Systems with Applications, 39 (2012), 11487-11493.

Krohling, R.A., Campanharo, V.C. Fuzzy TOPSIS for group decision making: A case study for accidents with oil spill in the sea, Expert Systems with Applications 38(4) (2011) 4190-4197.

Krohling, R.A., Pacheco, A.G.C., & Siviero, A.L.T. IF-TODIM: An intuitionistic fuzzy TODIM for decision making. Knowledge-Based Systems, 53 (2013), 142-146.

Lourenzutti, R.A., & Krohling, R.A. A study of TODIM in a Intuitionistic fuzzy and random environment. Expert Systems with Applications, 40, (2013), 6459-6468.

Lourenzutti, R. & Krohling, R.A. The Hellinger distance in multi-criteria decision making: An illustration to the TOPSIS and TODIM methods. Expert Systems with Applications, 41 (2014) 4414-4421.

Lourenzutti, R.A., & Krohling, R.A. A generalized TOPSIS method for group decision making with heterogeneous information in a dynamic environment. Information Sciences 330(10) (2016), 1-18.

Mahdavi, I., Mahdavi-Amiri, N., Heidarzade, A., Nourifar, R. Designing a model of fuzzy TOPSIS in multiple criteria decision making. Applied Mathematics and Computation, 206(2) (2008), 607-617.

Park, J. H., Park, I.Y., Kwun, Y.C., Tan, X. Extension of the TOPSIS method for decision making problems under interval-valued intuitionistic fuzzy environment, Applied Mathematical Modelling 35(5), (2011) 2544-2556.

Patel, P., Rahimi, S., Khorasani, E. Applied Z-numbers, 2015 Annual Conference of the North American of Fuzzy Information Processing Society (NAFIPS). pp. 1-6, 2015.

Wang, J. Q., & Wang, R. Q. Hybrid Random multi-criteria decision-making approach with incomplete certain information. Chinese Control and Decision Conference (CCDC), 2008, 1444-1448.

Wang, J., Liu, S.Y., & Zhang, J. An extension of TOPSIS for fuzzy MCDM based on vague set theory. Journal of Systems Science and Systems Engineering, (2005) 14, 73-84.

Wang, T.-C. and Lee, H.-D. Developing a fuzzy TOPSIS approach based on subjective weights and objective weights. Expert Systems with Applications, 36 (2009), 8980-8985.

Wang, T.-C., Lee, H.-D. Developing a fuzzy TOPSIS approach based on subjective weights and objective weights, Expert Systems with Applications 36 (2009) 8980-8985.

Wang, Y.J., Lee, H.S., Lin, K. Fuzzy TOPSIS for multi-criteria decision making, International Mathematical Journal 3 (2003) 367-379.

Wentao, X., Huan, Q. A extended TOPSIS method for the stochastic multi-criteria decision making problem through interval estimation, In Proceedings of the 2nd International Workshop on Intelligent Systems and Applications (ISA) 2010, pp. 1-4.

Xiao, Z.-Q., Application of Z-numbers in multi-criteria decision making. IEEE 2014 International Conference on Informative and Cybernetics for Computational Social Systems (ICCSS), pp 1-6, 2014.

Ye, F. An extended TOPSIS method with interval-valued intuitionistic fuzzy numbers for virtual enterprise partner selection, Expert Systems with Applications 37(10) (2010) 7050-7055.

Yue, Z. TOPSIS-based group decision-making methodology in intuitionistic fuzzy setting, Information Sciences 277(1) (2014) 141-153.

Zadeh, L.A. A note on Z-numbers. Information Sciences, 181(14) (2011), 2923- 2932.

Zadeh, L.A. Fuzzy sets. Information and Control, 8 (1965), 338-353.

Zhang, X., Xu, Z. The TODIM analysis approach based on novel measured functions under hesitant fuzzy environment, Knowledge-Based Systems, 61 (2014), 48-58.

Zimmermann, H.J. Fuzzy set theory and its application. Boston: Kluwer Academic Publishers, 1991.